\pdfoutput=1

\documentclass[11pt]{article}
\usepackage{multirow}
\usepackage{graphicx}  
\usepackage{caption}
\usepackage{booktabs}
\usepackage{enumitem}
\usepackage{subcaption}
\usepackage{latexsym}
\usepackage{soul}
\usepackage{microtype}
\usepackage[]{acl}

\usepackage{times}
\usepackage{latexsym}

\usepackage[T1]{fontenc}

\usepackage[utf8]{inputenc}

\usepackage{microtype}
\usepackage{todonotes}
\usepackage{tabularx}
\usepackage{enumitem}
\usepackage{booktabs}
\usepackage[]{graphicx}
\usepackage{enumitem}

\title{APPDIA: A Discourse-aware Transformer-based Style Transfer Model for Offensive Social Media Conversations}

\author{Katherine Atwell\textsuperscript{*}, Sabit Hassan\textsuperscript{*}, Malihe Alikhani\\
  Computer Science Department, School of Computing and Information\\
  University of Pittsburgh, Pittsburgh, PA\\
  \texttt{\{kaa139,sah259,malihe\}@pitt.edu}\\}
\begin{document}

\maketitle

\begin{abstract}
Using style-transfer models to reduce offensiveness of social media comments can help foster a more inclusive environment. However, there are no sizable datasets that contain offensive texts and their inoffensive counterparts, and fine-tuning pretrained models with limited labeled data can lead to the loss of original meaning in the style-transferred text. To address this issue, we provide two major contributions. First, we release the first publicly-available, parallel corpus of offensive Reddit comments and their style-transferred counterparts annotated by expert sociolinguists. Then, we introduce the first \textit{discourse-aware} style-transfer models that can effectively reduce offensiveness in Reddit text while preserving the meaning of the original text. These models are the first to examine inferential links between the comment and the text it is replying to when transferring the style of offensive Reddit text. 
We propose two different methods of integrating discourse relations with pretrained transformer models and evaluate them on our dataset of offensive comments from Reddit and their inoffensive counterparts. Improvements over the baseline with respect to both automatic metrics and human evaluation indicate that our discourse-aware models are better at preserving meaning in style-transferred text when compared to the state-of-the-art discourse-agnostic models.
\end{abstract}
\section{Introduction}
\textit{Disclaimer: Due to the nature of this work, figures and examples may contain offensive phrases.}

The spread of offensive and hateful content on social media can be detrimental to users' psychological well-being \cite{waldron2012harm,gulaccti2010effect}. Anonymity on platforms such as Reddit can further embolden users to post such hateful content \cite{ascher2019unmasking}. 
Further, the sheer volume of content on popular social media platforms can render the human moderation process ineffective \cite{hassan2022studying} or psychologically damaging for moderators \cite{10.1145/3290605.3300372} and calls for AI systems that can mitigate this problem.

AI moderation of social media by simply removing content classified as offensive \cite{zampieri-etal-2020-semeval,hassan-etal-2021-asad} may reduce diversity in online conversations and deter users from using the platform \cite{10.1145/3359294}. Our exploration reveals that many comments removed by moderators on Reddit contain contributions to the discourse beyond their offensive content. Rather than simply removing these comments from social media platforms, they can be turned into inoffensive statements by using alternative words, removing profanity, or paraphrasing certain parts, while preserving the overall semantic content. 

\setlength{\tabcolsep}{4pt}

\begin{figure}
    \centering
    \includegraphics[width=\linewidth]{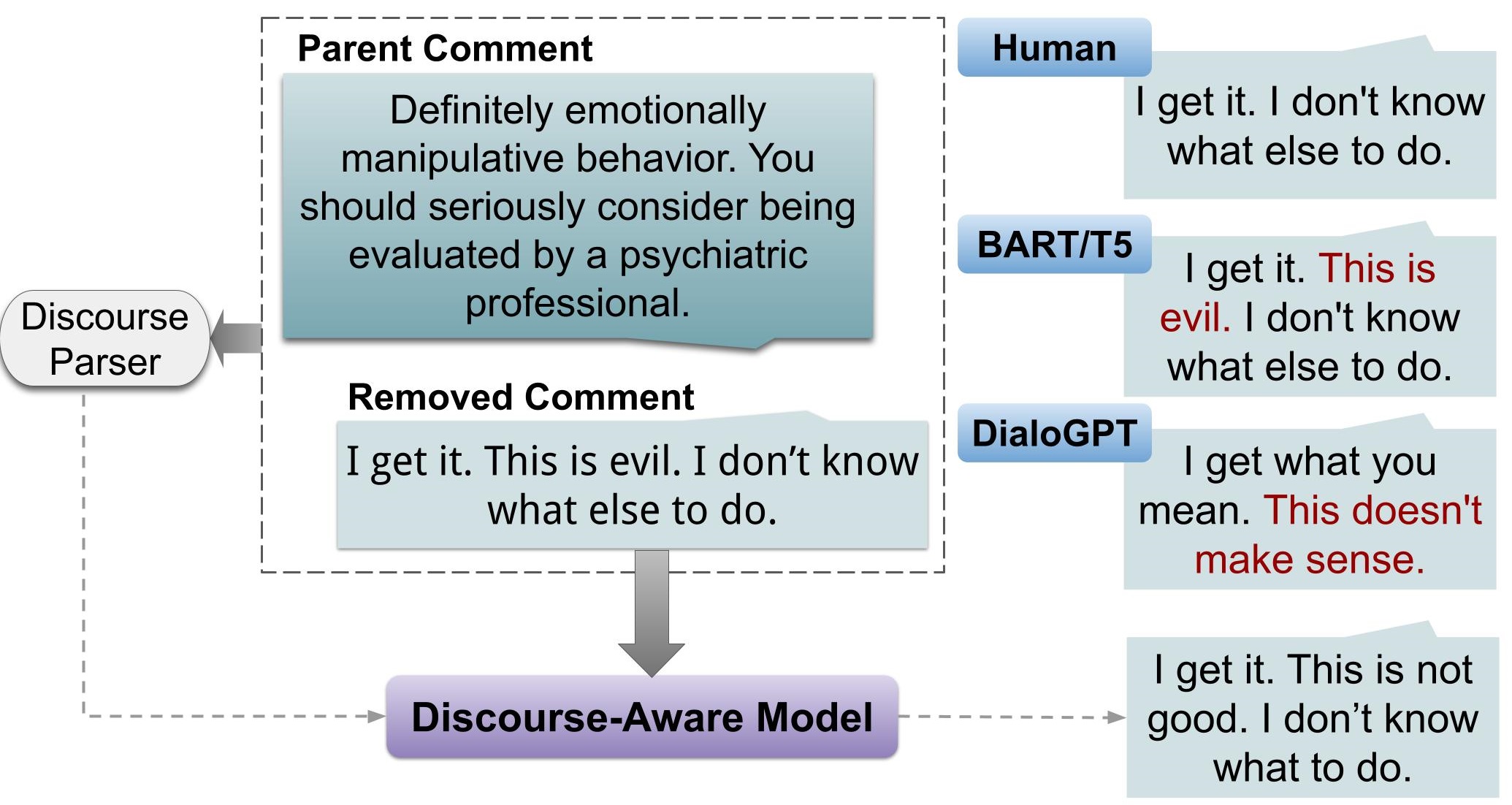}
    \caption{Example of instances where pretrained language models either fail to remove offensiveness (BART/T5) or drastically alter the intended meaning (DialoGPT) when fine-tuned on our style transfer task}
    \label{fig:intro_fig}
\end{figure}

We approach the problem of eliminating offensiveness from text while preserving original semantic content as a supervised \textit{style-transfer} task, where offensive text is transferred to inoffensive text. As a first step towards this goal, we create the first publicly-available, expert-annotated style transfer corpus for Reddit data, which contains offensive comments that include certain lexical items and more subtle instances that are implicit and grounded in context.  
This differentiates our work from unsupervised approaches are mostly good at handling instances with explicit lexical cues \cite{nogueira-dos-santos-etal-2018-fighting}.  

Although large pretrained transformer models have been successfully deployed for generation tasks, these models come with the risk of either failing to generate desired output or obfuscating the source passage's meaning while still producing coherent text \cite{10.1145/3442188.3445922}. 
In our work, we target the issue of content preservation using \emph{discourse frameworks}, which have been successfully employed for various generation tasks \cite{maskharashvili-etal-2021-neural, xu2022we, bosselut-etal-2018-discourse}, but have not been employed in style transfer models.
We hypothesize that integrating discourse coherence frameworks within transformer-based style transfer models
can contribute to better preservation of semantic content, specifically for short social media comments. 

We study our hypothesis with a small pilot annotation of style-transferred text produced by pretrained transformer models. Figure \ref{fig:intro_fig} shows examples of the issues described above in our style transfer task, where BART \cite{Lewis2020BARTDS} and T5 \cite{Raffel2020ExploringTL} do not remove offensiveness from the original comment, but DialoGPT \cite{Zhang2020DIALOGPTL} significantly alters the original semantic content.  We observe that coherence relations between a comment and its reply are not preserved under style transfer in cases where offensiveness is removed. For instance, the removed comment refers to \textit{"emotionally manipulative behavior"} in the parent comment with \textit{"This is evil"}, exhibiting the behavior of \textit{"Same-Unit"} discourse relation, which is not preserved in the style-transferred text generated by DialoGPT. 

To test our hypothesis, we provide the following contributions:

\begin{itemize}[leftmargin=*]
    \item Collect a dataset\footnote{ \url{https://github.com/sabithsn/APPDIA-Discourse-Style-Transfer}}     
 of \textasciitilde2K offensive comments from Reddit that are annotated by expert sociolinguists with inoffensive counterparts. Our data also contains parent comments/posts and are tagged with discourse relations, making it the first publicly available dataset of its kind.
    \item Propose two approaches for integrating discourse relation frameworks with pretrained transformer models:
    i) using Penn Discourse Treebank \cite{miltsakaki-etal-2004-penn, prasad2008penn, webber2019penn} relations within a single comment, and ii) parsing a comment and the text it is responding to using the Rhetorical Structure Theory Discourse Treebank \cite{Mann1988RhetoricalST}.
    
    
\end{itemize}

The results for both discourse-aware approaches indicate improvement in content preservation over the pretrained baselines, providing support for our hypothesis and for the use of discourse frameworks to preserve meaning in style-transfer tasks. 

\section{Related Work}

Paraphrase generation is a well-studied problem that has yielded large datasets such as the PDTB paraphrase database \cite{Ganitkevitch2013PPDBTP}, WikiAnswer \cite{Fader2013ParaphraseDrivenLF}, ParaNMT \cite{Wieting2018ParaNMT50MPT}, and the MSCOCO dataset \cite{Lin2014MicrosoftCC}. Recent works in the related but relatively new field of style transfer primarily target sentiment transfer \cite{li-etal-2018-delete,yu-etal-2021-rethinking-sentiment}, formality transfer \cite{chawla-yang-2020-semi} or expertise transfer \cite{cao-etal-2020-expertise}. Very few works have targeted transferring style from offensive to inoffensive text, with \citet{nogueira-dos-santos-etal-2018-fighting} and \citet{cheng-etal-2020-contextual} being notable exceptions. Our dataset differs from the aforementioned works in multiple ways. Ours is the first publicly available dataset that contains offensive Reddit comments that are rewritten by experts, paired with parent comment/post, and automatically tagged with discourse relations. Further, both \citet{nogueira-dos-santos-etal-2018-fighting} and \citet{cheng-etal-2020-contextual} derive their datasets from political subreddits \cite{Serban2017ADR}, while our data encompasses subreddits on personal views, question-answer discussions, and gender rights in addition to political subreddits. 
\begin{figure*}[]
    \centering
    \includegraphics[width=.69\textwidth]{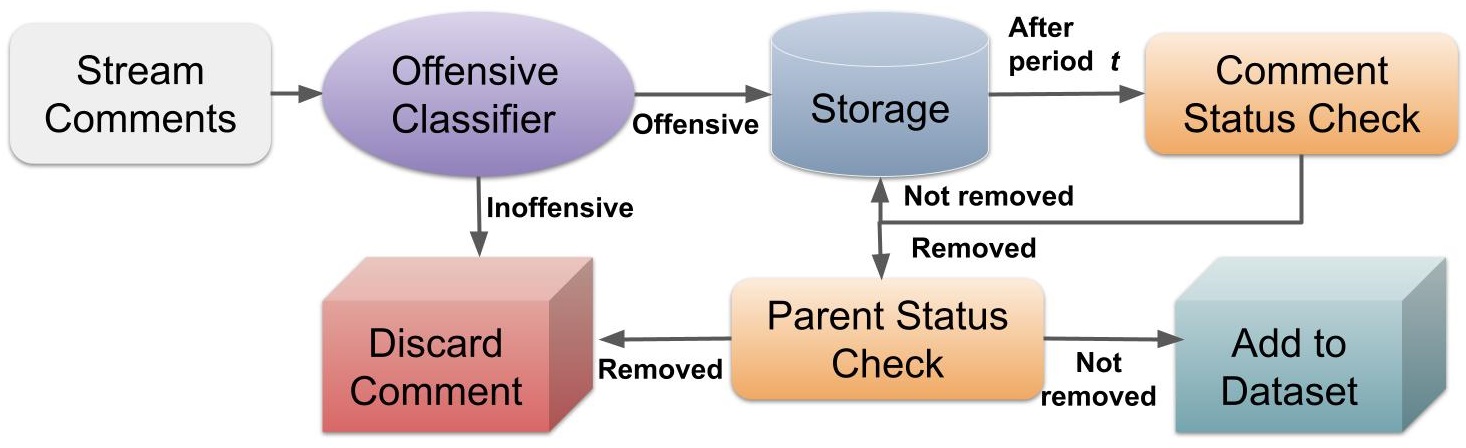}
    \caption{Our data collection pipeline for obtaining removed comments from Reddit that are offensive.}
    \label{fig:data_collection_pipeline}
\end{figure*}

Development of pretrained language models (PLM) such as BART \cite{Lewis2020BARTDS} has changed the landscape of natural language generation research and we are witnessing a shift toward controllable text generation \cite{Zhang2022ASO,Zeldes2020TechnicalRA,Ribeiro2021StructuralAI,Ziegler2019FineTuningLM}. Discourse relations have been proposed as a possible controlled generation method, shown to aid extractive and abstractive summarization \cite{Cohan2018ADA,Xu2020DiscourseAwareNE}, text generation from meaning representations \cite{maskharashvili-etal-2021-neural}, and question answering with logical reasoning or complex answers \cite{huang2021dagn, xu2022we}. Discourse-aware models have also been shown to generate more coherent texts \cite{bosselut-etal-2018-discourse} within a reinforcement learning setting. 
Our work integrates both RST-DT and PDTB frameworks with pretrained transformer models and provides a comparison of the relative efficacy of the two frameworks for a generation task. Within the context of style transfer, recent works have focused on classification and reconstruction loss \cite{nogueira-dos-santos-etal-2018-fighting,chawla-yang-2020-semi} in semi-supervised/unsupervised setting, use of copy mechanism \cite{Jhamtani2017ShakespearizingML}, and coherence classifier \cite{cheng-etal-2020-contextual} to guide the style transferred text. To our knowledge, our work is the first to utilize discourse coherence frameworks for style transfer.

\section{Data Collection and Annotation}
In order to reduce offensiveness in text, we create an expert-annotated dataset of offensive comments and their style-transferred counterparts. In this section, we first describe our pipeline for collecting and curating a set of offensive comments from Reddit. Then, we describe our annotation process for reducing offensiveness in these comments.

\subsection{Data Collection Pipeline}
First, we stream 14 subreddits spanning topics of politics, personal views, question-answer discussions, and gender rights for new comments using PRAW\footnote{https://praw.readthedocs.io/en/stable/}. The streamed comments are then tagged for offensiveness using a BERT model \cite{Devlin2019BERTPO} fine-tuned on the OLID dataset \cite{Zampieri2019PredictingTT}, which consists of 14K tweets annotated for offensiveness and was used for the SemEval 2019 \cite{Zampieri2019SemEval2019T6} shared task. If a comment is tagged as inoffensive by the classifier, we remove it from our data. As our initial exploration revealed that a large portion of the removed comments on Reddit (> 60\%) may not be offensive and may have been removed due to violation of subreddit-specific rules, the exclusion of inoffensive comments from the data is essential for a feasible annotation process. Our manual annotation, as described in the next section, eliminates any false positive bias that the classifier may have. More details about the classifier can be found in the Appendix.

After running the classifier, the body and metadata of comments that are tagged as offensive, are stored locally. We then periodically check the accessibility status of these comments on Reddit. If it has been removed by moderators, we query Reddit for the \emph{parent} comment or post that it is in response to. If the comment is a reply to another comment, then the comment it is replying to is considered the parent, and if the comment is a top-level comment, i.e, a direct reply to a post, then the post is considered the parent. If the parent has been deleted or removed, the comment is discarded. Otherwise, we add the comment, along with the parent, to our dataset. Our data collection pipeline is summarized in Figure \ref{fig:data_collection_pipeline}. After filtering out very long comments, we end up with a pool of 5.8K comments for annotation. 



\renewcommand{\arraystretch}{1.2}%

\begin{table*}[]
    \small
    \centering
    \begin{tabular}{p{5cm}|p{4cm}|p{2cm}|p{3.5cm}}
    \textbf{Original Comment} & \textbf{Rewritten Comment} & \textbf{Global/ Local} & \textbf{Reason for paraphrasing} \\
    \hline
    \hline
    You can't do s*** because you're an idiot. & You can't do anything because you're not competent. & Local & Cursing, Insults \\
    \hline
    So you s**k as person. Got it & So you're not a great person. Got it & Local & Cursing, Insults \\
    \hline
    What backward b*****k nowhere country do you live in? & What country do you live in? & Local & Xenophobia, Cursing \\
    \hline
    Keep my phone gallery secrets out your f***** MOUTH & Don't talk about my phone gallery secrets & Global & Cursing, Rudeness\\
    \hline
    F*** off.  Sick of people like you thinking  everything is propaganda & Please go away. Tired of people like you thinking everything has a hidden plan & Global & Cursing, Rudeness \\
    \hline
    To hell with peaceful protest. Protesters should drag DeathSantis out of his home and have a public trial & Peaceful protest won't work. Protesters should go for a public trial & Global & Threats of Violence \\
    \hline

    \hline
    \end{tabular}
         \caption{Examples of applying local and global changes to the comments for different types of offensive speech, as per our annotation protocol.} 

    \label{tab:annotation-examples}
\end{table*}
\subsection{Data Annotation}
The 5.8K comments obtained from our data collection pipeline are annotated by three expert sociolinguists. The annotators are paid 30 USD per hour and the Institutional Review Board (IRB) protocol, as defined by our institution, the University of Pittsburgh, was followed for recruitment and annotation. 

The primary goal of our annotation is to remove offensiveness from a comment while retaining the intent of the comment.  Similar to the SemEval 2019 shared task \cite{Zampieri2019SemEval2019T6}, we define offensiveness as consisting of insults, profane words, hate speech, or threats of violence for our purposes. We observe that some comments can be made inoffensive by the removal or substitution of offensive words. We call such changes \textit{localized changes}. For other comments, however, the text needs to be altered/paraphrased substantially to reduce offensiveness. We refer to this type of change as \textit{global change}. With these principles in mind, the annotators are provided with an annotation protocol, whose key points are listed below:
\begin{itemize}[leftmargin=*,itemsep=1pt]
    \item Each comment has to be manually inspected. If a comment is already inoffensive, or cannot be translated into inoffensive text without altering the original intent, it is discarded.
    \item If applying \textit{localized changes} is not possible or doesn't rid the comment of offensiveness, then \textit{global changes} are made. 
\end{itemize}
Examples of our manual annotation can be found in Table \ref{tab:annotation-examples}. The first three rows of Table \ref{tab:annotation-examples} show examples of localized changes and the last three rows show examples of globalized changes in our data. Further details about the distribution of data and subreddits can be found in Appendix \ref{A:data}.

To assess the meaning preservation of annotation, we compute the BLEU score \cite{Papineni2002BleuAM} between the annotated text and the original text. We use the BLEU score to measure similarity due to the open-endedness of the task (the inter-rater agreement, for instance, cannot be calculated here). Since BLEU compares the overlap between reference and candidate sentences, it can serve as a metric for measuring content preservation. Our annotations achieve a BLEU score of \textbf{60.06} with the original text as reference. Since a BLEU score of 60 generally indicates a high overlap with the reference sentences, we can deduce that our annotation process successfully preserved the original meaning. Further, the offensiveness classifier is used to tag the annotated text, showing that annotators eliminated offensiveness from \textbf{68\%} of the comments. In reality, however, this number is likely to be higher, as the classifier may tag inoffensive comments about sensitive subjects as offensive. For example, "a rape victim should not be the one to blame" is tagged as offensive. This highlights the limitations of existing offensiveness classifiers.
\section{Discourse-Aware Models}
We propose two approaches for integrating the PDTB and RST-DT discourse frameworks into pretrained transformer models, as described below.

\begin{figure}[!h]
\centering
\includegraphics[width=\linewidth]{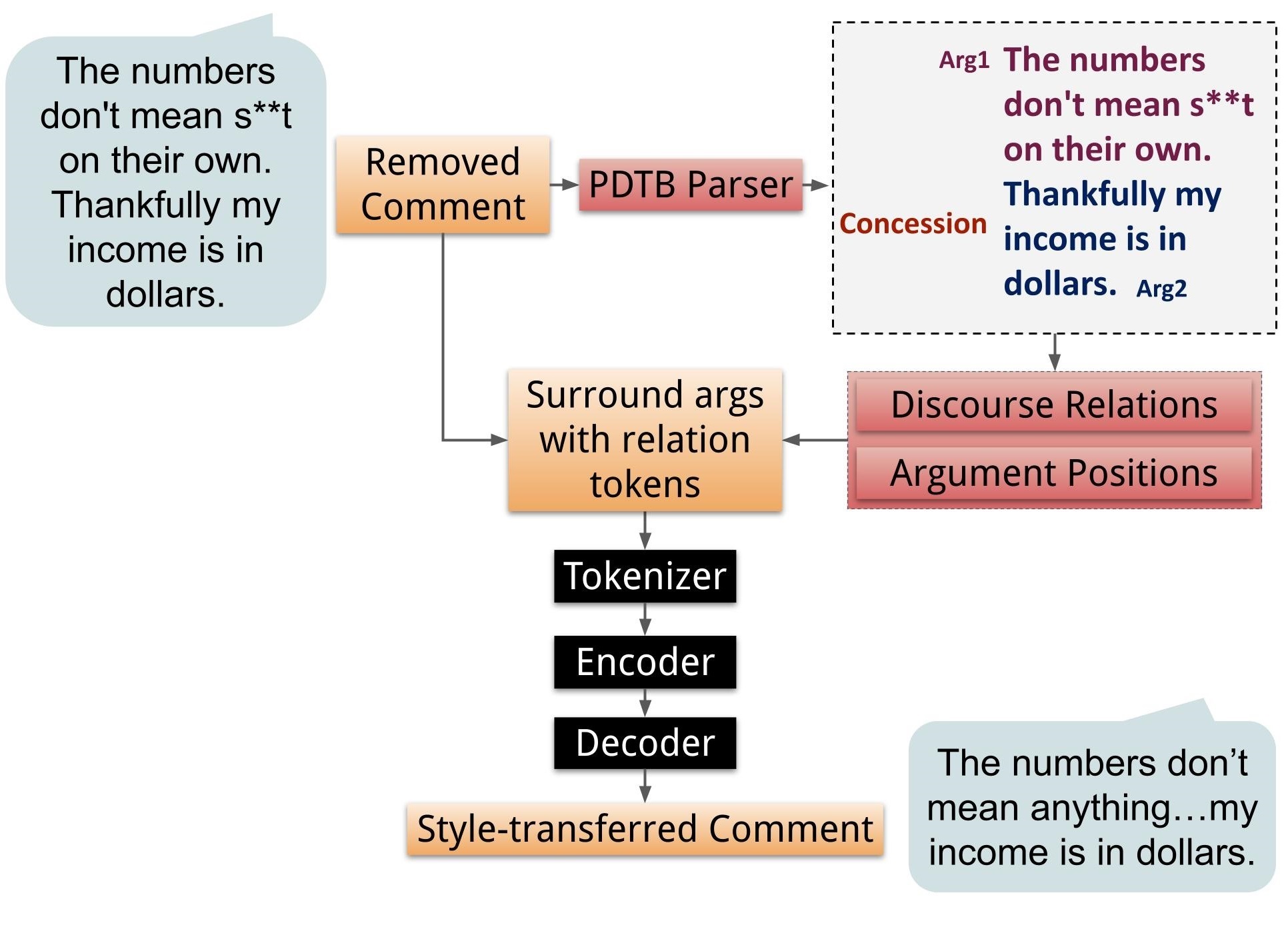}
\caption{PDTB-augmented style transfer model. Special tokens represent the beginning and end of each argument, as well as the relation between each argument pair, are passed to the encoder.}
\label{fig:pdtbmodel}
\end{figure}

\begin{figure}[!h]
\centering
\includegraphics[width=\linewidth]{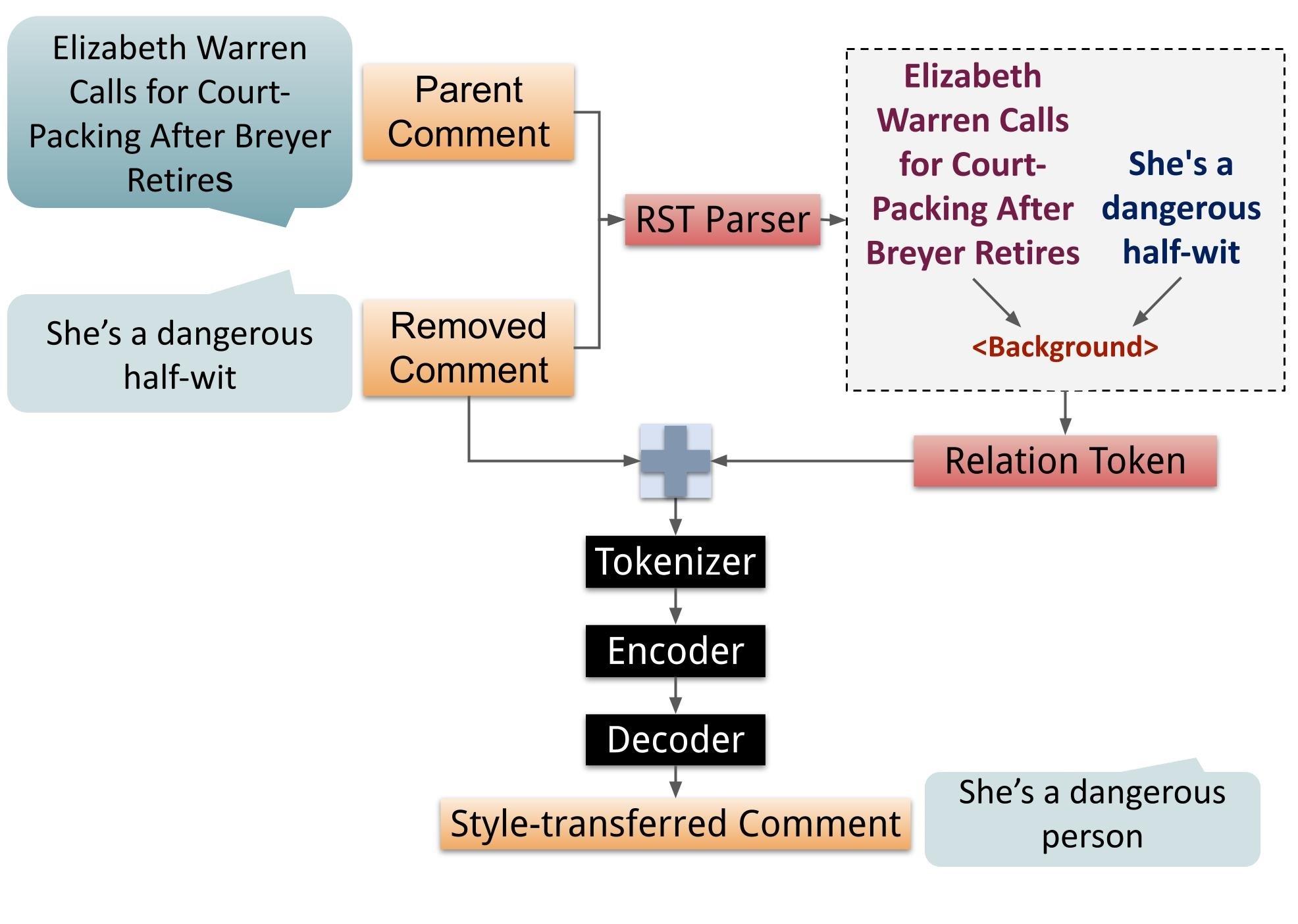}
\caption{RST-augmented style transfer model. A special token representing the relation at the root of the RST tree is prepended to the tokenized text of the removed comment, which is then passed to the encoder.}
\label{fig:rstmodel}
\end{figure}

\paragraph{PDTB Within-Comment Relations}
To extract PDTB relations at the comment level, we parse the comment text in isolation, first using the \citet{lin2014pdtb} end-to-end discourse parser to extract explicit discourse relations (signaled by a discourse connective), then running the \citet{kim-etal-2020-implicit} XLNet-large model to extract implicit discourse relations (not signaled by a discourse connective) from adjacent sentence pairs. Because there is no PDTB-tagged Reddit corpus available, we run these models trained on the PDTB-2 corpus. For the L2 classification task (the more difficult of the tasks we run), \citet{lin2014pdtb} report an F-1 score of 80.61, and \citet{kim-etal-2020-implicit} report an accuracy of 57.74 (they do not report F-1 for the L2 classification task) on the PDTB-2. We then use the positions of the argument pairs, and their discourse relations, in our input.

\paragraph{RST-DT Context-Based Relations}

To obtain a representation of the RST-DT relation between a comment and its parent, we concatenate the contents of the comment and the parent, separating them out as paragraphs. We then run the \citet{ijcai2018-579} EDU segmenter on this text, and run the model in \citet{wang2017two} on the resulting EDUs. We train and test this parser on the RST-DT and GUM corpus \citet{10.1007/s10579-016-9343-x} combined, and report the F-1 scores on the test set in Appendix \ref{sec:appendix}. We use the relation at the root of the RST tree as input to our style-transfer model. 

\paragraph{Integration with transformer model} To integrate the RST-DT and PDTB relations within pretrained transformer models, we first generate special tokens representing each relation for RST-DT and for the start and end of each relation for PDTB. We update the tokenizer with these additional tokens, insert the tokens in the input text, and pass the modified text to the encoder of the model, as shown in Figures \ref{fig:pdtbmodel} (PDTB) and \ref{fig:rstmodel} (RST-DT). We resize the model embedding to accommodate for this additional vocabulary.

\section{Experiments}
In this section, we first describe the experiments with pretrained transformer models, followed by the experiments with discourse-aware models.
\subsection{Pretrained Transformer Models}
We experiment with three different pretrained transformer models, namely: i) BART-base \cite{Lewis2020BARTDS}, ii) T5-base \cite{Raffel2020ExploringTL}, and iii) DialoGPT-medium \cite{Zhang2020DIALOGPTL}. While BART and T5 are pretrained on formal data such as Wikipedia or web data such as C4\footnote{https://www.tensorflow.org/datasets/catalog/c4}, DialoGPT is pretrained on Reddit data for the response generation task. 

\subsection{Discourse-aware Transformer Models}
Due to its higher potential in removing offensiveness, we integrate our discourse-aware approaches with DialoGPT. To integrate PDTB relations, we experiment with the following  variations: i) Level 1 and Level 2 explicit PDTB relations, ii) Level 1 and Level 2 implicit PDTB relations, and iii) combining level 2 explicit and implicit relations. To incorporate RST-DT, we use our proposed approach with the top-level RST-DT classes. We limit our scope to only top-level RST-DT classes because we are unlikely to encounter lower-level classes frequently in our dataset. 

We also experiment with combining both of our approaches. Under this setting, a comment is prepended with root-level RST-DT relation between itself and its parent, and PDTB relations (both implicit and explicit) are inserted in the body of the text. Since PDTB implicit and RST parsers have low accuracy scores, we propose setting a threshold \textbf{$\alpha$} for the inclusion of a discourse relation. If the confidence score for a given relation falls below $\alpha$, the relation is discarded. This is done to account for higher likelihood of misclassification on instances the discourse classifiers have low confidence on. We experiment on three different $\alpha$ values as follows: 

\begin{enumerate}
    \item We set $\alpha=0$, and thus all predicted RST-DT and PDTB relations are taken
    \item We compute the mean ($\mu$) and standard deviation ($\sigma$) of the classifier score for the predicted class and set $\alpha = \mu-\sigma$
    \item We compute the interquartile range of the classifier scores and set $\alpha = Q1$, where $Q1$ is the value of first quartile.
\end{enumerate}

\section {Results}
Below, we describe the results of our experiments. We split our dataset into an 80-10-10 split for training, development, and test sets respectively. We first calculate automatic metrics, reporting the BLEU \cite{Papineni2002BleuAM} and rescaled BERTScore \cite{Zhang2020BERTScoreET} on our test set. In addition, we compute the \textit{SafeScore} \textemdash percentage of style transferred comments predicted as \textit{inoffensive} by the BERT classifier that was initially used to identify potential candidates. Further, we ask a human annotator to compare style-transferred text generated by baseline model and our proposed discourse-aware model.

\subsection{Automatic Evaluation}

Using our automated metrics, we compare semantic similiarity between: i) the manual annotation and style transferred text, and ii) the original comment and style-transferred text. 

\subsubsection{Pretrained Models}
While BART and T5 are seen to achieve very high BLEU and BERT scores in Table \ref{tab:baseline-models}, these numbers hide critical failures of the models: staying too close to the original comment and not reducing offensiveness. The goal of an ideal style transfer model would be to have a good SafeScore, while also achieving a good BLEU and BERTScore. A good point of reference for this ideal scenario would be the BLEU, BERTScore, and SafeScore achieved by human annotators. DialoGPT, in contrast to BART and T5, has a lower BLEU and BERTScore, but is notably better at reducing offensiveness and achieves SafeScore comparable to that of human annotators. This could be attributed to the fact that unlike BART and T5, which are pretrained on out-of-domain web or formal data, DialoGPT is specifically pretrained on Reddit data, making it suitable for our task. For the rest of the paper, we refer to DialoGPT as the baseline model.

\begin{table}[h]
    \small
    \centering
    \begin{tabular}{l|c|c||c}
    \hline
    \multicolumn{4}{c}{\textbf{Compared Against Annotated Text}}\\
    \hline
    \textbf{Model} & \textbf{BLEU} & \textbf{BERTScore} & \textbf{SafeScore}\\
    \hline
    BART & 65.1	& 68.1 & 44.7\\
    \hline
    T5 & 65.3 & 69.2 & 51.3\\    
    \hline
    DialoGPT & 42.5	& 47.2 & 66.3\\
    \hline
    \multicolumn{4}{c}{\textbf{Compared Against Original Text}}\\
    \hline
    \textbf{Model} & \textbf{BLEU} & \textbf{BERTScore} & \textbf{SafeScore}\\
    \hline
    BART & 76.2	& 78.4 & 44.7\\
    \hline
    T5 & 74.8 & 78.0 & 51.3\\    
    \hline
    DialoGPT & 45.3	& 49.4 & 66.3\\
    \hline
    \end{tabular}
         \caption{Results of finetuning pretrained models on our dataset. While BART and T5 outputs have a high similarity to the original and annotated text, they do not drastically reduce offensiveness, while the reverse is true for DialoGPT.}

    \label{tab:baseline-models}
\end{table}
\renewcommand{\arraystretch}{1.2}%

\begin{table*}[!h]
    \small
    \centering
    \begin{tabular}{l|l|c|c||c}
    \hline
    \hline
    \multicolumn{1}{l}{\textbf{Discourse Framework}} & \multicolumn{1}{l}{\textbf{Discourse Relations}} & \multicolumn{1}{l}{\textbf{BLEU}} & \multicolumn{1}{l}{\textbf{BERTScore}} & \multicolumn{1}{l}{\textbf{SafeScore}}\\
        \hline
    \hline
    \multicolumn{5}{c}{\textbf{Compared Against Annotated Text}}\\
    \hline
    \hline

    None (Baseline) & - & 42.5 (0.0)	& 47.2	(0.0) & 66.3 (0.0)\\

    \hline
    & Lvl 1 - Explicit & 42.6 (0.0) & 46.5 (-0.7) & 63.3 (-3.0)\\
    \cline{2-5}
    
    & Lvl 1 - Implicit &  44.3	(1.8) & 48.9 (1.7)	& 65.8 (-0.5)\\
    \cline{2-5}
    
    PDTB ($\alpha$ = 0)& Lvl 2 - Explicit & 42.5	(0.0) & 47.1 (-0.1) & 64.3 (-2.0)\\
    \cline{2-5}
    & Lvl 2 - Implicit & 43.9 (1.3) & 48.9 (1.7) & 65.0 (-1.3)\\
    \cline{2-5}
    & Lvl 2 - Explicit + Implicit & 44.4 (1.8)	& 48.7	(1.5) & 65.3 (-1.0)\\
    \hline
    RST ($\alpha$ = 0)& Top-level & 45.2 (2.6)	& \textbf{50.6 (3.4)} & 65.7 (-0.7)\\    
    \hline
    RST + PDTB ($\alpha$ = 0)& Lvl 2 - Explicit + Implicit (PDTB), Top-level (RST) & \textbf{46.7 (4.2)}	& 50.3 (3.1) &	\textbf{67.7 (1.3)}\\
    RST + PDTB ($\alpha$ = $\mu-\sigma$)& Lvl 2 - Explicit + Implicit (PDTB), Top-level (RST) & 46.5 (4.0)	& \textbf{50.6 (3.4)} &	66.0 (-0.3)\\ 
    RST + PDTB ($\alpha$ = $Q1$)& Lvl 2 - Explicit + Implicit (PDTB), Top-level (RST) & 45.6 (3.1)	& 50.2	(3.0) & 64.3 (-2.0)\\ 
    \hline
    \hline

    \multicolumn{5}{c}{\textbf{Compared Against Original Text}}\\
    \hline
        \hline

    None (Baseline) & - & 45.3	(0.0) & 49.4 (0.0) & 66.3 (0.0)\\
    \hline
        & Lvl 1 - Explicit & 46.1 (0.8) & 49.0 (-0.4) & 63.3 (-3.0)\\
    \cline{2-5}
    & Lvl 1 - Implicit & 46.7 (1.4)	& 50.3	(0.9) &  65.8 (-0.5)\\
    \cline{2-5}
    
    PDTB ($\alpha$ = 0) & Lvl 2 - Explicit & 46.2	(0.0) & 49.6 (0.2) & 63.5 (-2.8) \\
    \cline{2-5}
    & Lvl 2 - Implicit & 46.9 (1.6) & 51.0 (1.6) & 65.0 (-1.3) \\
    \cline{2-5}
    & Lvl 2 - Explicit + Implicit & 47.2 (1.9) & 50.8 (1.4) & 65.3 (-1.0)\\
    \hline
    RST ($\alpha$ = 0) & Top-level & 47.9 (2.5) & \textbf{52.8 (3.4)} & 65.7 (-0.7)\\    
    \hline
    RST + PDTB ($\alpha$ = 0) & Lvl 2 - Explicit + Implicit (PDTB), Top-level (RST)& \textbf{49.6	(4.3)} & 52.6 (3.2) &	\textbf{67.7 (1.3)}\\
    RST + PDTB ($\alpha$ = $\mu-\sigma$)& Lvl 2 - Explicit + Implicit (PDTB), Top-level (RST) & 49.4 (4.1) &	51.5 (2.0) &	66.0 (-0.3)\\ 
    RST + PDTB ($\alpha$ = $Q1$)& Lvl 2 - Explicit + Implicit (PDTB), Top-level (RST) & 47.8 (2.5) & 51.5 (2.0) & 64.3 (-2.0)\\ 
    \hline
    \hline

    \end{tabular}
         \caption{Results from running our discourse-aware style transfer models, where the average numbers across three runs are reported and the best numbers for each metric are bolded. Improvement over baseline is shown in parenthesis. As the above tables demonstrate, incorporating discourse relations improves model results by a wide margin, with RST root-level relations yielding the best BERTScore results and the combined RST + PDTB model yielding the best offensiveness score and BLEU score results.}

    \label{tab:discourse-models}
\end{table*}

\begin{table*}[!h]
    \small
    \centering
    \begin{tabular}{l|l}
    \hline
   \textbf{Original Comment} &\textbf{Really bad stance. What an unbelievable moron you are.}\\
    \hline
    Manual Rewrite & This is a really bad stance.\\
    \hline
    Baseline & That is a bad stance, you are\\
    \hline
    Discourse-aware & That's a bad stance, and you are being unbelievable.\\
    \hline
    \hline
   \textbf{Original Comment} &\textbf{Just like Republicans creating themselves narratives to fuel their persecution complex to }\\
      &\textbf{justify their hatred for everyone that isn't on the Jesus c***}\\
    \hline
    Manual Rewrite & Just like Republicans creating themselves narratives as fuel to justify their hatred for everyone  \\
    & that isn't on their side.\\
    \hline
    Baseline & Just like Democrats creating themselves a narrative to justify themselves's hatred for \\
    & people that aren't on their side.\\
    \hline
    Discourse-aware & Republicans create their own narratives to justify themselves.\\
    \hline
    \hline
       \textbf{ Original Comment} & \textbf{Youre soft as baby s***}\\
    \hline
    Manual Rewrite & Youre really soft\\
    \hline
    Baseline & Youre being so rude\\
    \hline
    Discourse-aware & Youre quite soft\\
    \hline
    \hline
       \textbf{ Original Comment} & \textbf{Yeah that dude is a punk a**}\\
    \hline
    Manual Rewrite & Yeah that dude is awful\\
    \hline
    Baseline & Yeah that guy is not cool.\\
    \hline
    Discourse-aware & Yeah that dude has issues \\
    \hline
    \hline
       \textbf{Original Comment} & \textbf{This is so cringe. What a f*****g loser.}\\
    \hline
    Manual Rewrite & This is so cringe.\\
    \hline
    Baseline & This is not cool at all.\\
    \hline
    Discourse-aware & This is so sad. Such a loser. \\
    \hline
   
    \hline
    \end{tabular}
         \caption{Examples of style-transferred text generated by the different models. The discourse-aware model refers to our best-performing discourse-aware model, the RST-PDTB model $(\alpha=0)$. The top three examples are ones in which our model performed better than the baseline, while in the fourth example both performed well and in the bottom example the baseline performed better than the discourse-aware model.}

    \label{tab:generation-examples}
\end{table*}

\subsubsection{Discourse-Aware Models}
Table \ref{tab:discourse-models} shows improvement in automated metrics achieved by our discourse aware models in comparison to the baseline DialoGPT, providing strong evidence in favor of our hypothesis. In addition to this overarching takeaway, we make the following observations from our experiments:

\paragraph{The choice of framework impacts performance} Although discourse models yield improvements on the baseline for each automatic metric, the extent of improvement over the baseline varies depending on the discourse framework used. Most notably, the RST-DT relation between the comment and its parent has the highest individual impact on BLEU and BERTScore, suggesting that the context of a comment is important for models to retain semantic meaning in generated text. While we do not see any major difference between Level 1 and Level 2 PDTB relations, implicit PDTB relations have a higher impact on the BLEU and BERTScore than explicit PDTB relations. Although implicit relation parsers have a lower accuracy, the improvement can be attributed to the fact that implicit relations occur more frequently in our dataset (41\% instances) compared to explicit relations that occur in 25\% of the instances. Further, explicit relations are lexically signalled by discourse connectives already present in the text, while implicit relations do not have connectives present in the text. Combining implicit and explicit relations does not change the performance notably. 

\paragraph{Combining discourse frameworks yields the highest improvement} Combining our approaches for PDTB and RST-DT relations has the greatest impact on the BLEU score, with an absolute improvement of \textbf{4.3} over the baseline. The BLEU score, in this case, is a measure of overlap with the original content, while the SafeScore measures the efficacy of offensiveness removal. The better BLEU score with the highest SafeScore of \textbf{67.7} indicates that incorporating both discourse frameworks enables the model to preserve original content better while effectively removing offensiveness compared to other approaches. Although the BERTScore is slightly lower than that achieved by RST-augmented model, the improvement of \textbf{3.2} over baseline supports the use of both frameworks. 

\paragraph{Low-confidence relations are important}
Our last observation is that dropping low-confidence relations ($\alpha=\mu - \sigma$) can negatively impact SafeScore, while BLEU and BERTscore remains relatively unchanged. We notice that, if value of $\alpha$ is increased ($\alpha=Q1$), then the BLEU and BERTScore begin to degrade. This suggests, while classifier accuracy is a concern for implicit PDTB and RST-DT relations, the classifiers still capture valuable information that can aid the preservation of semantic content and reduction of offensiveness.

\subsection{Human Evaluation}

Although automated metrics such as BLEU and BertScore can be good indicators of preservation of original content, they have certain limitations. For example, they do not take into account cases where deviating from the original comment is the correct approach for offensiveness reduction. We also observe that, in certain cases, the models may generate text that has a high overlap in words but their coherence may be affected by out-of-place words. Thus, human evaluation is required for a complete understanding of limitations and strengths of our proposed model. 

To this end, we presented one of our expert annotators with 100 randomly selected examples and style transferred text generated by both the baseline and our best discourse-aware model. The order of the text generated by the two models was randomly shuffled so that the human evaluation was free from any potential bias. Table \ref{tab:generation-examples} shows examples of style-transferred text generated by the different models. The expert annotator was asked to judge each pair from three angles: i) which of the generated texts preserves the original semantic content most, ii) which of the generated texts is more coherent, and iii) which of the generated text is preferred overall. 

We report the results of the human evaluation in two different dimensions. First, we analyze all 100 samples to get an overall picture of improvement. Next, we exclude comments that do not contain any discourse relation. This allows us to understand how much effect discourse relations may have on the overall results. From the evaluation results reported in Table \ref{tab:human-eval}, we make the following key observations described below.

\begin{table}[h]
    \small
    \centering
    \begin{tabular}{l|c|c||c}
    \multicolumn{4}{c}{\textbf{Full human evaluation set}}\\
    \hline
    \textbf{Preferred Model} & \textbf{Content-} & \textbf{Coherence} & \textbf{Overall}\\
    & \textbf{Preservation} & &\\
    \hline
    Baseline & 36\% & 32\% & 29 \% \\
    \hline
    Discourse-aware & \textbf{48\%} & \textbf{37\% }& \textbf{40\%}\\    
    \hline
    No preference& 16\% & 31\% & 31\%\\
    \hline
    \multicolumn{4}{c}{\textbf{Subset with discourse relations}}\\
    \hline
    \textbf{Preferred Model} & \textbf{Content-} & \textbf{Coherence} & \textbf{Overall}\\
    & \textbf{Preservation} & &\\
    \hline
    Baseline & 30\% & 34\% & 26 \% \\
    \hline
    Discourse-aware & \textbf{56\%} & \textbf{46\% }& \textbf{46\%}\\    
    \hline
    No preference& 14\% & 20\% & 28\%\\
    \hline
    
    \hline
    \end{tabular}
         \caption{Results of human evaluation}

    \label{tab:human-eval}
\end{table}
\paragraph {Discourse improves both coherence and content preservation} While we see a large preference for our discourse-aware model overall (\textbf{40\%} as opposed to 29\%), the difference is more prominent in terms of content preservation (\textbf{48\%} vs 36\%) compared to coherence (\textbf{37\%} vs 32\%). This further supports our hypothesis that, while the baseline model can generate coherent texts, a discourse-aware model is necessary for content preservation.
\paragraph{Improvements are larger for comments containing discourse relations} For the subset of data where discourse relations are present, we see an even larger improvement of our discourse model compared to the baseline. Our model is preferred in \textbf{56\%} of cases for content preservation (compared to 30\% for the baseline), \textbf{46\%} for coherence (compared to the baseline's 34\%) and \textbf{46\%} overall (compared to 26\% for the baseline). This implies that the difference between our model and the baseline becomes more important for comments that have discourse structure within them. 





\renewcommand{\arraystretch}{1.2}%

\section{Conclusion and Future Work}

In this paper, we have demonstrated that utilizing discourse frameworks and parsing models can help pretrained transformer models preserve original content when transferring style from offensive to inoffensive. We have shown that combining different discourse frameworks can further improve content preservations. The improvements we observe in this paper are significant; however, we hypothesize that utilizing discourse relations for these tasks can be even more impactful if the performance of existing discourse parsers is improved. Discourse parsing is a very challenging task \cite{atwell-etal-2021-discourse,atwell-etal-2022-change}, and the largest and most widely-used corpora are composed of news texts over a short time span. Thus, there is a need for further research (and additional annotated corpora) on discourse relations within the context of social media. We hope our publicly available code and data will motivate other researchers to build on the groundwork laid out in this paper. 

Further research is also necessary in the context of style-transferring for offensive text. After further improving these language models and evaluating their safety,
future systems that are proven to be robust and effective can potentially help social media moderators or be deployed in a human-in-the-loop or assistive technology capacity. We expect these models to have the potential to not only improve the psychological well-being of users but also to motivate healthy engagement on social media. 

\section{Ethical Considerations}
We acknowledge that our models can not eliminate offensiveness completely from a given text. Thus, deploying our model to display style-transferred text requires taking future safety measures. We also acknowledge that our use of pretrained models can induce biases in certain scenarios, as pretrained models have been shown to be susceptible to bias in the data used for pretraining \cite{Li2021Robustness}.

\section*{Acknowledgement}
We would like to thank SRI International for their valuable feedback. This project was supported by DARPA grant prime OTA No. HR00112290024 (subcontract No. AWD00005100 and SRA00002145). We also
acknowledge the Center for Research Computing at the University of Pittsburgh for providing computational resources. We would also like to thank the human annotators, the anonymous reviewers, Ilana Torres, and Mert Inan for their valuable feedback.

\bibliography{anthology,custom}
\bibliographystyle{acl_natbib}

\clearpage
\appendix
\newpage

\section{Data Annotation}
\label{A:data}
\begin{table}[!h]
    \small
    \centering
    \begin{tabular}{|c|c|c|}
    \hline
    \textbf{Group} & \textbf{Subreddits}  & \textbf{Counts}\\
    \hline
	& r/Conservative	& 457\\
	& r/PoliticalCompassMemes	& 69\\
	& r/politics	& 241\\
Politics & r/PoliticalHumor	& 315\\
	& r/conspiracy & 167\\
	& r/socialism	& 21\\
	& r/Anarcho\_Capitalism & 29\\
	\hline
\multicolumn{2}{|l}{\textbf{Subtotal}} &  \textbf{1299}\\
\hline
		
	& r/unpopularopinion & 181\\
Personal & r/ChangeMyView & 131\\
    views & r/AmITheAsshole & 73\\
	& r/offmychest & 81\\
	\hline
	\multicolumn{2}{|l}{\textbf{Subtotal}} &  \textbf{466}\\
	\hline
	& r/AskReddit & 66\\
Question- & r/askscience & 11\\
	Answer& r/AskHistorians	 & 7\\
	& r/explainlikeimfive	& 95\\
	\hline
	\multicolumn{2}{|l}{\textbf{Subtotal}} &  \textbf{179}\\
    \hline
Gender	& r/MensRights	& 23\\
	Rights & r/FemaleDatingStrategy &	14\\
	\hline
	\multicolumn{2}{|l}{\textbf{Subtotal}} &  \textbf{37}\\
    \hline
    \hline
\multicolumn{2}{|l}{\textbf{Total}} &  \textbf{1981}\\
\hline
    \end{tabular}
         \caption{Distribution of annotated data}

    \label{tab:distribution}
\end{table}

\paragraph{Annotation Distribution} Following the annotation process, we obtain a labeled set of \textasciitilde2K comments with their corresponding rewrites. Table \ref{tab:distribution} shows the distribution of the annotated data. From this distribution, we observe that frequency of offensive comments are high in political subreddits such as \textit{r/Conservative} compared to popular subreddits such as \textit{r/AskReddit}. Subreddits such as \textit{r/MensRights} did not yield a substantial number of rewrites. Analyzing our data revealed two reasons for the low frequency: i) the traffic on these subs is low compared to other subreddits, and ii) removed comments from these subreddits frequently contain extremely toxic content that cannot be rewritten into non-offensive versions while preserving original intent. These particular subreddits need to be streamed for a longer period to obtain a substantial number of offensive comments that can be rewritten as non-offensive.

\section {Pretrained Model Hyperparameters}
\paragraph{Offensiveness classifier:} We fine-tune bert-base-cased \cite{Devlin2019BERTPO} for 3 epochs on the OLID training set \cite{Zampieri2019PredictingTT}. We use learning rate of 8e-5, batch size of 8 and maximum length of 100. The model achieved an F1 score of 80.2 on the OLID test set.
\paragraph{Style transfer models:} For all style transfer models, we use the same set of hyperparameters: block size of 512, batch size of 2, learning rate of 5e-5. All models were fine-tuned for 10 epochs. During generation, we again use set of parameters: maximum length of 200, top\_p of 0.7 and temperature of 0.8.

\section{Performance of RST parser}

\label{sec:appendix}
\begin{table}[h]
\small
\centering
\begin{tabular}{c|c}
\toprule
\textbf{Relation} & \textbf{F1} \\
\midrule
Attribution          & 0.8214 \\
Background           & 0.2121 \\
Cause                & 0.0769 \\
Comparison           & 0.0870 \\
Condition            & 0.5714 \\
Contrast             & 0.3059 \\
Elaboration          & 0.4753 \\
Enablement           & 0.5263 \\
Evaluation           & 0.0000 \\
Explanation          & 0.1728 \\
Joint                & 0.3769 \\
Manner-Means         & 0.3636 \\
Same-Unit            & 0.7417 \\
Summary              & 0.3704 \\
Temporal             & 0.1047 \\
Textual-Organization & 0.2105 \\
Topic-Change         & 0.0250 \\
Topic-Comment        & 0.0000 \\
span                 & 0.6656
\end{tabular}
\caption{F-1 scores for RST parser trained on RST and GUM data and tested on an evaluation set from each (details in text)}
\label{discourse_parser_f1}
\end{table}

\end{document}